\documentclass{article}
\usepackage[T1]{fontenc} 
\usepackage[utf8]{inputenc} 
\usepackage{ismir,amsmath,cite,url}
\usepackage{graphicx}
\usepackage{color}
\usepackage{tikz}
\usepackage{caption}  
\usepackage{spverbatim}

\title{Learning to Generate Music With Sentiment}

\twoauthors
 {Lucas N. Ferreira} {University of California, Santa Cruz \\ Department of Computational Media}
 {Jim Whitehead} {University of California, Santa Cruz \\ Department of Computational Media}

\begin{document}

\maketitle
\begin{abstract}
Deep Learning models have shown very promising results in automatically composing polyphonic music pieces. However, it is very hard to control such 
models in order to guide the compositions towards a desired goal. 
We are interested in controlling a model to automatically 
generate music with a given sentiment. This paper presents a 
generative Deep Learning model that can be directed to compose music 
with a given sentiment. Besides music generation, the same model can be used 
for sentiment analysis of symbolic music. We evaluate the accuracy of the model 
in classifying sentiment of symbolic music using a new dataset of video game 
soundtracks. Results show that our model is able to obtain good prediction
accuracy.  A user study shows that human subjects agreed that the generated music has the intended sentiment, however negative pieces can be ambiguous.
\end{abstract}
\section{Introduction}
\label{sec:introduction}

Music Generation is an important application domain of Deep Learning 
in which models learn musical features from a dataset in order to
generate new, interesting music. Such models have been capable of generating high 
quality pieces of different styles with strong short-term dependencies\footnote{
Supporting strong long-term dependencies (music form) is still an open problem.}\cite{briot2017deep}. 
A major challenge of this domain consists of disentangling these 
models to generate compositions with given characteristics. For example, 
one can't easily control a model trained on classical piano pieces 
to compose a tense piece for a horror scene of a movie. Being able to 
control the output of the models is specially important for the field of
Affective Music Composition, whose major goal is to automatically 
generate music that is perceived to have a specific emotion or 
to evoke emotions in listeners \cite{williams2015investigating}. Applications involve generating 
soundtracks for movies and video-games \cite{williams2015dynamic}, sonification 
of biophysical data \cite{Chen2015} and generating responsive music 
for the purposes of music therapy and palliative care \cite{miranda2011brain}.

Recently, Radford et al. \cite{radford_2017} showed that a generative 
Long short-term memory (LSTM) neural network can learn an excellent 
representation of sentiment (positive-negative) on text, despite being 
trained only to predict the next character in the Amazon reviews 
dataset \cite{He2016}. When combined to a Logistic Regression, 
this LSTM achieves state-of-the-art sentiment analysis accuracy on the 
Stanford Sentiment Treebank dataset and can match 
the performance of previous supervised systems using 30-100x fewer 
labeled examples. This LSTM stores almost all of the 
sentiment signal in a distinct ``sentiment neuron'', which 
can be used to control the LSTM to generate sentences with a given
sentiment. In this paper, we explore this approach with the goal of composing 
symbolic music with a given sentiment. We also explore this approach as
a sentiment classifier for symbolic music.


In order to evaluate this approach, we need a 
dataset of music in symbolic format  
that is annotated by sentiment. Even though emotion detection 
is an important topic in music information retrieval \cite{kim2010music}, 
it is typically studied on music in audio format. To the best of 
our knowledge, there are no datasets of symbolic music annotated
according to sentiment. Therefore, we created a new dataset composed 
of 95 MIDI labelled piano pieces (966 phrases of 4 bars) from video game soundtracks.  
Each piece is annotated by 30 human subjects according to a valence-arousal (dimensional) 
model of emotion \cite{russell1980circumplex}. The sentiment of each piece is then 
extracted by summarizing the 30 annotations and mapping the valence axis to sentiment. The same dataset also contains another 728 non-labelled 
pieces, which were used for training the generative LSTM.


We combine this generative LSTM with a Logistic
Regression and analyse its sentiment prediction accuracy against a traditional classification LSTM trained in a fully-supervised way. Results showed that our model (generative LSTM with Logistic Regression) outperformed the supervised LSTM
by approximately 30\%. We also analysed the generative capabilities of our model with a user study.
Human subjects used an online annotation tool to label 3 pieces controlled to
be negative and 3 pieces controlled to be positive. Results showed human annotators agree the generated positive pieces have the intended sentiment. The generated negative pieces appear to be ambiguous, having both negative and positive parts.

We believe this paper is the first work to explore sentiment analysis in
symbolic music and it presents the first disentangled Deep Learning model
for music generation with sentiment. Another contribution of this paper
is a labelled dataset of symbolic music annotated according to sentiment.
These contributions open several direction for future research, 
specially music generation with emotions as both a multi-class 
problem and as a regression problem. Moreover, these methods
could be applied to create soundtrack generation systems 
for films, video games, interactive narratives, audio books, etc.

\section{Related Work}
\label{sec:related_work}

This paper is related to previous work on Affective Algorithmic Music Composition, 
more specifically to works that process music in symbolic form in order to generate music 
with a given emotion. A common approach for this problem consists of designing a rule-based 
system to map musical features to a given emotion in a categorical or dimensional space 
\cite{williams2015investigating}. For example, Williams et al. \cite{williams2015dynamic} 
propose a system to generate soundtracks for video games where each game's scene graph 
(defining all the possible branching of scenes in the game) is annotated 
according to a valence-arousal model. A second-order Markov model is used to learn 
melodies from a dataset and are then transformed by a rule-based system to fit 
the annotated emotions in the graph. Davis and Mohammad \cite{davis2014generating} 
follow a similar approach in TransPose, a system that composes piano melodies 
for novels. TransPose uses a lexicon-based approach to automatically detect emotions 
(categorical model) in novels and a rule-based technique to create piano melodies with 
these emotions.

There are a few other approaches in the literature to compose music with a given emotion. 
Scirea et al. \cite{scirea2017affective} recently presented a framework called MetaCompose designed 
to create background music for games in real-time. MetaCompose generates music
by (i) randomly creating a chord sequence from a pre-defined chord progression graph, (ii) 
evolving a melody for this chord sequence using a genetic algorithm and
(iii) producing an accompaniment for the melody/chord sequence combination.
Monteith et al. \cite{monteith2010automatic} approaches Affective Algorithmic Music Composition from a Machine Learning perspective to learn melodies and rhythms from a corpus of music 
labeled according to a categorical model of emotion. Individual Hidden Markov models and n-grams are trained for
each category to generate pitches and underlying harmonies, respectively. Rhythms 
are sampled randomly from examples of a given category.

Deep Learning models have recently achieved high-quality results in music 
composition with short-term dependencies \cite{briot2017deep}. These models normally are
trained on a corpus of MIDI files to predict the next note to be played based on a given note. In general, 
these models can't be manipulated to generate music with a given emotion. 
For example, in the system DeepBach, Hadjeres et al. \cite{hadjeres2017deepbach} 
use a dependency network and a Gibbs-like sampling procedure to generate high-quality 
four-part chorales in the style of Bach. Roberts et at. \cite{Roberts2017} train recurrent 
variational autoencoder (VAEs) to reproduce short musical sequences and with a novel hierarchical 
decoder they are able to model long sequences with musical structure for both individual
instruments and a three-piece band (lead, bass, and drums). 

The majority of the deep learning models are trained to generate musical scores and 
not performances. Oore et al. \cite{oore2017learning} tackles this problem by 
training an LSTM with a new representation that 
supports tempo and velocity events from MIDI files. This model was trained on the 
Yamaha e-Piano Competition \cite{yamahaEPiano}, which contains MIDI captures of \textasciitilde1400 
performances by skilled pianists. With this new representation and dataset,
Oore et al. \cite{oore2017learning} generated more human-like performances when compared to previous models. 

\section{Model}
\label{sec:model}

We propose a Deep Learning method for affective algorithmic composition that
can be controlled to generate music with a given sentiment. This method is based on the work of 
Radford et al. \cite{radford_2017} which generates product reviews (in textual form) with sentiment.
Radford et al. \cite{radford_2017} used a single-layer multiplicative 
long short-term memory (mLSTM) network \cite{krause2017} with 4096 units to process 
text as a sequence of UTF-8 encoded bytes (character-based language modeling). 
For each byte, the model updates its hidden state of the mLSTM and predicts a 
probability distribution over the next possible byte. 


    
    
    
    
    

This mLSTM was trained on the Amazon product review dataset, which contains over 82 million 
product reviews from May 1996 to July 2014 amounting to over 38 billion 
training bytes \cite{He2016}. Radford et al. \cite{radford_2017} used the trained mLSTM 
to encode sentences from four different Sentiment Analysis datasets.
The encoding is performed by initializing the the states to zeros and 
processing the sequence character-by-character. The final hidden states of 
the mLSTM are used as a feature representation. With the encoded
datasets, Radford et al. \cite{radford_2017} trained a simple logistic 
regression classifier with L1 regularization and outperformed 
the state-of-the-art methods at the time using 30-100x fewer 
labeled examples.

By inspecting the relative contributions of features on various datasets, 
Radford et al. \cite{radford_2017} discovered a single unit within the 
mLSTM that directly corresponded to sentiment. Because the mLSTM was 
trained as a generative model, one can simply set the value of the 
sentiment unit to be positive or negative and the model generates 
corresponding positive or negative reviews.

\subsection{Data Representation}
\label{sec:representation}

We use the same combination of mLSTM and logistic
regression to compose music with sentiment. To do this, we
treat the music composition problem as a language modeling problem.
Instead of characters, we represent a music piece as a 
sequence of words and punctuation marks from a vocabulary that represents 
events retrieved from the MIDI file. Sentiment is perceived in 
music due to several features such as melody, harmony, tempo, timbre, 
etc \cite{kim2010music}. Our data representation attempts to encode a large part 
of these features\footnote{Constrained by the features one can extract from MIDI data.} 
using a small set of words:

\begin{itemize}
    \item ``n\_[pitch]'': play note with given pitch number: any integer from 0 to 127.
    \item ``d\_[duration]\_[dots]'': change the duration of the following notes to a given 
    duration type with a given amount of dots. Types are breve, whole, half, quarter, 
    eighth, 16th and 32nd. Dots can be any integer from 0 to 3.
    \item ``v\_[velocity]'': change the velocity of the following  notes to a given velocity (loudness) number. Velocity is discretized in 
    bins of size 4, so it can be any integer in the set $V = {4, 8, 12, \dots, 128}$.
    \item ``t\_[tempo]'': change the tempo of the piece to a given tempo in bpm. Tempo is also discretized in bins of size 4, so it can be any integer in the set $T = {24, 28, 32, \dots, 160}$.
    \item ``.'': end of time step. Each time step is one sixteenth note long.
    \item ``\textbackslash n'': end of piece.
\end{itemize}

\begin{figure}
 \centerline{
 \includegraphics[width=\columnwidth]{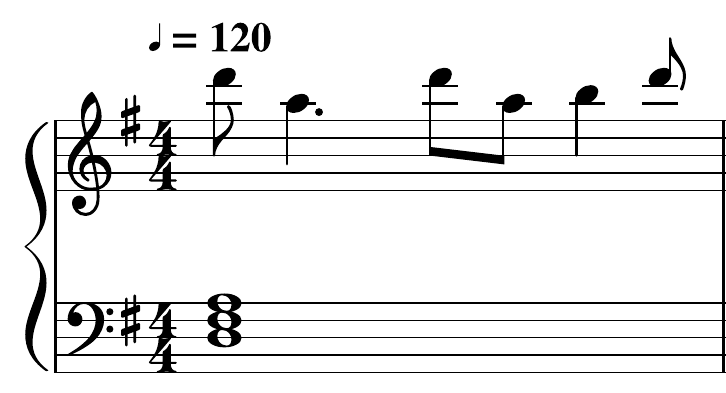}}
\begin{spverbatim}
t_120 v_76 d_whole_0 n_50 n_54 n_57
v_92 d_eighth n_86 . . v_84 d_quarter_1 n_81 . .
\end{spverbatim}
 
 \caption{A short example piece encoded using our proposed representation. The encoding represents the first two time steps of the shown measure.}
 \label{fig:enc_ex}
\end{figure}

For example, Figure \ref{fig:enc_ex} shows the encoding of the first two time steps of the 
first measure of the Legend of Zelda - Ocarina of Time's Prelude of Light. 
The first time step sets the tempo to 120bpm, the velocity of the following notes to 
76 and plays the D Major Triad for the duration of a whole note. The second time step sets the velocity to 84 and plays a dotted quarter A5 note. The total size of this vocabulary is 225 and it represents both the composition and performance elements of a piece (timing and dynamics). 

\section{Sentiment Dataset}
\label{sec:dataset}

In order to apply the Radford et al. \cite{radford_2017} method to compose music with 
sentiment, we also need a dataset of MIDI files to train the LSTM and another one
to train the logistic regression. There are many good datasets of music in MIDI format 
in the literature. However, to the best of our knowledge, none are labelled according to sentiment. Thus, we created a new 
dataset called VGMIDI which is composed of 823 pieces extracted
from video game soundtracks in MIDI format. We choose video game soundtracks 
because they are normally composed to keep the player in a certain affective state 
and thus they are less subjective pieces. All the pieces are piano 
arrangements of the soundtracks and they vary in length  from 26 seconds to 3 minutes. 
Among these pieces, 95 are annotated according to a 2-dimensional model
that represents emotion using a valence-arousal pair. Valence indicates 
positive versus negative emotion, and arousal indicates emotional intensity \cite{Soleymani_2013}.

We use this valence-arousal model because it allows continuous annotation of music 
and because of its flexibility---one can directly map a valence-arousal (v-a) pair to 
a multiclass (happy, sad, surprise, etc) or a binary (positive/negative) 
model. Thus, the same set of labelled data permits the investigation of affective algorithmic music composition as both a classification 
(multiclass and/or binary) and as a regression problem. The valence-arousal model
is also one of the most common dimensional models used to label emotion in music 
\cite{Soleymani_2013}.

Annotating a piece according to the v-a model consists of continuously 
listening to the piece and deciding what valence-arousal pair best represents the emotion 
of that piece in each moment, producing a time-series of v-a pairs. 
This task is subjective, hence there is no single ``correct'' time-series for a given
piece. Thus, we decided to label the pieces by asking several human subjects
to listen to the pieces and then considering the average time-series as the ground truth. 
This process was conducted online via Amazon Mechanical Turk, where each 
piece was annotated by 30 subjects using a web-based tool we designed 
specifically for this task. Each subject annotated 2 pieces out of 95,
and got rewarded USD \$0.50 for performing this task.

\subsection{Annotation Tool and Data Collection}
\label{sec:data_collection}

The tool we designed to annotate the video game soundtracks in MIDI format 
is composed of five steps, each one being a single web-page. 
These steps are
based on the methodology proposed by Soleymani et al. \cite{Soleymani_2013} for annotating 
music pieces in audio waveform. First, participants are introduced to the annotation 
task with a short description explaining the goal of the task and how long it 
should take in average. Second, they are presented to the definitions of valence and 
arousal. In the same page, they are asked to play two short pieces and indicate 
whether arousal and valence are increasing or decreasing. Moreover, we ask the 
annotators to write two to three sentences describing the short pieces they 
listened to. This page is intended to measure their understanding of the valence-arousal model 
and willingness to perform the task.  Third, a video tutorial was 
made available to the annotators explaining how to use the 
annotation tool. Fourth, annotators are exposed to the main 
annotation page.

\begin{figure}
 \centerline{
 \includegraphics[width=\columnwidth]{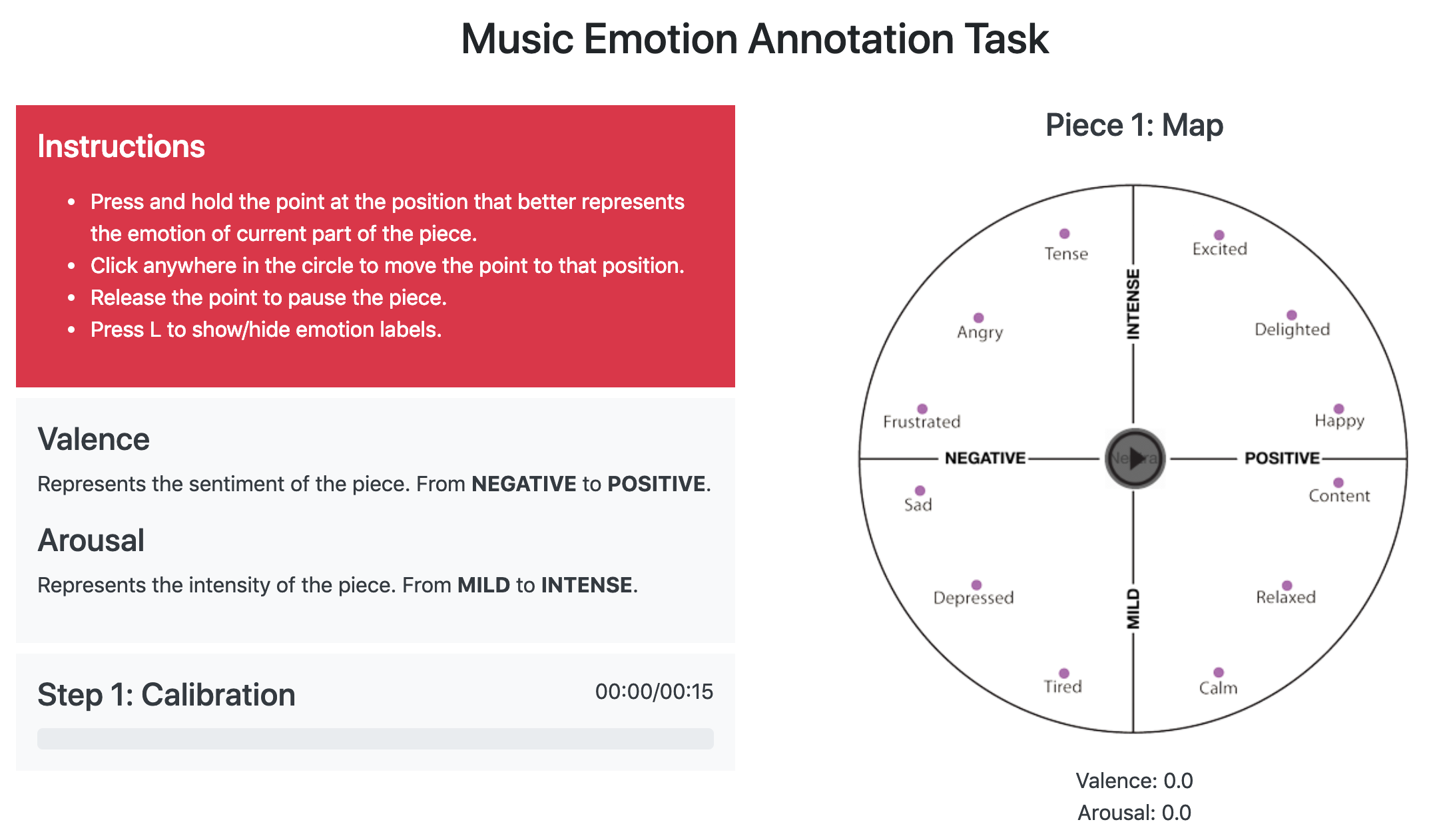}}
 \caption{Screenshot of the annotation tool.}
 \label{fig:annotation_main}
\end{figure}

This main page has two phases: calibration and annotation.
In the calibration phase, annotators listen to the first 15 seconds of the
piece in order to get used to it and to define the starting
point of the annotation circle. In the annotation phase they 
listen to the piece from beginning to end and label it using the annotation circle, which
starts at the point defined during the calibration phase.
Figure \ref{fig:annotation_main} shows the annotation interface 
for valence and arousal, where annotators click and hold the circle (with the play icon) 
inside the v-a model (outer circle) indicating the current emotion 
of the piece. In order to maximize annotators’ engagement in the task, the piece 
is only played while they maintain a click on the play circle.  
In addition, basic instructions on how to use the tool are showed to the participants
along with the definitions of valence and arousal. A progression bar is also 
showed to the annotators so they know how far they are from completing each phase. 
This last step (calibration and annotation) is repeated for a second piece.
All of the pieces the annotators listened to are MIDI files synthesized with the 
``Yamaha C5 Grand" soundfont. Finally, after the main annotation step, participants 
provide demographic information including gender, age, location (country), musicianship 
experience and whether they previously knew the pieces they annotated. 

\subsection{Data Analysis}
\label{sec:data_analysys}

The annotation task was performed by 1425 annotators, where 55\% are female and 42\%
are male. The other 3\% classified themselves as transgender female, 
transgender male, genderqueer or choose not to disclose their gender. 
All annotators are from the United States and have an average age of approximately 31 years. 
Musicianship experience was assessed using a 5-point Likert scale where 1 means 
``I've never studied music theory or practice'' and 5 means ``I have an undergraduate 
degree in music''. The average musicianship experience is 2.28. They spent on average 
12 minutes and 6 seconds to annotate the 2 pieces.

The data collection process provides a time series of valence-arousal values for
each piece, however to create a music sentiment dataset we only need the valence 
dimension, which encodes negative and positive sentiment. 
Thus, we consider that each piece has 30 time-series of valence values. The 
annotation of each piece was preprocessed,
summarized into one time-series and split into ``phrases'' 
of same sentiment. The preprocessing is intended to remove 
noise caused by subjects performing the 
task randomly to get the reward as fast as possible. 
The data was preprocessed by smoothing each annotation with moving average
and clustering all 30 time-series into 3 clusters (positive, negative and noise) 
according to the dynamic time-warping distance metric. 

\begin{figure}
 \centerline{
 \includegraphics[width=\columnwidth]{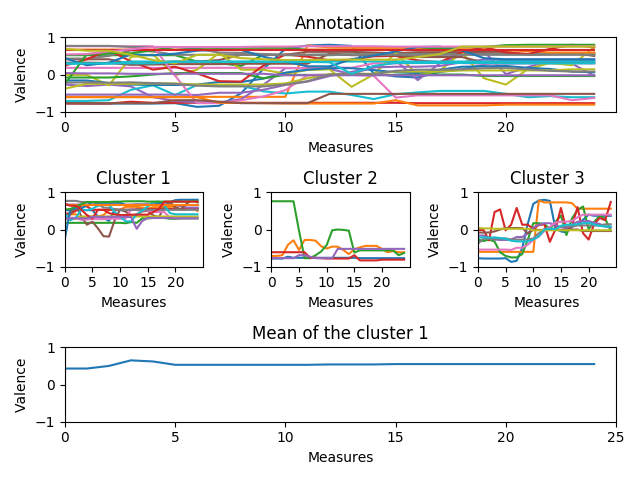}}
 \caption{Data analysis process used to define the final label of the phrases of a piece. }
 \label{fig:clustering}
\end{figure}

We consider the cluster with the highest variance to be noise cluster and so we 
discard it. The cluster with more time series among the two remaining ones is 
then selected and summarized by the mean of its time series. 
We split this mean into several segments with the same sentiment. 
This is performed by splitting the 
mean at all the points where the valence changes 
from positive to negative or vice-versa. Thus,
all chunks with negative valence are considered phrases 
with negative sentiment and the ones with positive valence are positive 
phrases. Figure \ref{fig:clustering} shows an example of this three-steps process 
performed on a piece. All the phrases that had no notes (i.e. silence phrases)
were removed. This process created a total of 966 phrases: 599 positive 
and 367 negative. 


\section{Sentiment Analysis Evaluation}

To evaluate the sentiment classification accuracy of our method (generative mLSTM + 
logistic regression), we compare it to a baseline method which is a 
traditional classification mLSTM trained in a supervised way. Our method uses 
unlabelled MIDI pieces to train a generative mLSTM to predict the next word in a 
sequence. An additional logistic regression uses the hidden states of the generative 
mLSTM to encode the labelled MIDI phrases and then predict sentiment. 
The baseline method uses only labelled MIDI phrases to train a 
classification mLSTM to predict the sentiment for the phrase. 

The unlabelled pieces used to train the generative mLSTM  were 
transformed in order to create additional training examples, 
following the methodology of Oore et al. 
\cite{oore2017learning}. The transformations consist of time-stretching (making 
each piece up to 5\% faster or slower) and transposition (raising or lowering 
the pitch of each piece by up to a major third). We then encoded all these 
pieces and transformations according to our word-based representation (see 
Section \ref{sec:representation}). Finally, the encoded pieces were shuffled 
and 90\% of them were used for training and 10\% for testing. The training set was divided into 3 shards of similar size (approximately 18500 pieces each -- 325MB) and the testing set was combined into 1 shard (approximately 5800 pieces -- 95MB). 

We trained the generative mLSTM with 6 different sizes (number of neurons
in the mLSTM layer): 128, 256, 512, 1024, 2048 and 4096. For each size, 
the generative mLSTM was trained for 4 epochs using the 3 training 
shards. Weights were updated with the Adam optimizer after processing 
sequences of 256 words on mini-batches of size 32. The mLSTM hidden 
and cell states were initialized to zero at the beginning of each 
shard. They were also persisted across updates to simulate 
full-backpropagation and allow for the forward propagation of 
information outside of a given sequence \cite{radford_2017}. Each sequence is 
processed by an embedding layer (which is trained together with 
the mLSTM layer) with 64 neurons before passing through the mLSTM layer. 
The learning rate was set to $5*10^{-6}$ at the beginning and decayed linearly 
(after each epoch) to zero over the course of training. 

We evaluated each variation of the generative mLSTM with a 
forward pass on 
test shard using mini-batches of size 32. Table 
\ref{tab:gen_anal} shows the average\footnote{Each mini-batch 
reports one loss.} cross entropy loss 
for each variation of the generative mLSTM.

\begin{table}[!h]
 \begin{center}
 \begin{tabular}{cc}
  \hline
  \textbf{mLSTM Neurons} & \textbf{Average Cross Entropy Loss}\\ \hline
  128 & 1.80   \\
  256  & 1.61  \\
  512  & 1.41  \\
  1024 & 1.25  \\
  2048 & 1.15  \\
  4096 & 1.11  \\ \hline
 \end{tabular}
\end{center}
\caption{Average cross entropy loss of the generative mLSTM with different amount of neurons.}
 \label{tab:gen_anal}
\end{table}

The average cross entropy loss decreases as the size of the mLSTM increases, reaching the best result (loss 1.11) when size is equal to 4096. Thus, we used the variation with 4096 neurons to proceed with the sentiment classification experiments. 

Following the methodology of Radford et al. \cite{radford_2017}, we re-encoded 
each of the 966 labelled phrases using the final cell states (a 4096 dimension vector) 
of the trained generative mLSTM-4096. The states are calculated by initializing them to zero 
and processing the phrase word-by-word. We plug a logistic regression into the mLSTM-4096 
to turn it into a sentiment classifier. This logistic regression model was trained with 
regularization ``L1'' to shrink the least important of the 4096 feature weights to zero. 
This ends up highlighting the generative mLSTM neurons that contain most of the sentiment
signal.

We compared this generative mLSTM + logistic regression approach against 
our baseline, the supervised mLSTM. This is an mLSTM with exactly the same architecture 
and size of the generative version, but trained in a fully supervised way.  
To train this supervised mLSTM, we used the word-based representation of the phrases, but we 
padded each phrase with silence (the symbol ``.'') in order to equalize their length. 
Training parameters (learning rate and decay, epochs, batch size, etc) were the same ones 
of the the generative mLSTM. It is important to notice that in this case the mini-batches are 
formed of 32 labelled phrases and not words. We evaluate both methods using a 10-fold cross 
validation approach, where the test folds have no phrases that appear in the training folds.  
Table \ref{tab:sent_anal} shows the sentiment classification accuracy of both approaches.

\begin{table}[!h]
 \begin{center}
 \begin{tabular}{lc}
  \hline
  \textbf{Method} & \textbf{Test Accuracy} \\ \hline
  Gen. mLSTM-4096 + Log. Reg. & 89.83 $\pm$ 3.14\\
  Sup. mLSTM-4096             & 60.35 $\pm$ 3.52 \\
  \hline
 \end{tabular}
\end{center}
\caption{Average (10-fold cross validation) sentiment classification accuracy of both generative (with logistic regression) and supervised mLSTMs.}
 \label{tab:sent_anal}
\end{table}
 
The generative mLSTM with logistic regression achieved an 
accuracy of 89.83\%, outperforming the supervised mLSTM by 29.48\%. The supervised
mLSTM  accuracy of 60.35\% suggests that the amount of labelled data (966 phrases) 
was not enough to learn a good mapping between phrases and sentiment.
The accuracy of our method shows that the generative mLSTM is capable of learning, in 
an unsupervised way, a good representation of sentiment in symbolic music.

This is an important result, for two reasons. First, since the higher accuracy of generative 
mLSTM is derived from using unlabeled data, it will be easier to improve this over time using additional 
(less expensive) unlabeled data, instead of the supervised mLSTM approach which requires additional (expensive) 
labeled data. Second, because the generative mLSTM was trained to predict the next word in a sequence, it can 
be used as a music generator. Since it is combined with a sentiment predictor, it opens up the possibility of 
generating music consistent with a desired sentiment. We explore this idea in the following section.

\section{Generative Evaluation}

\begin{figure}[!h]
 \centerline{
 \includegraphics[width=\columnwidth]{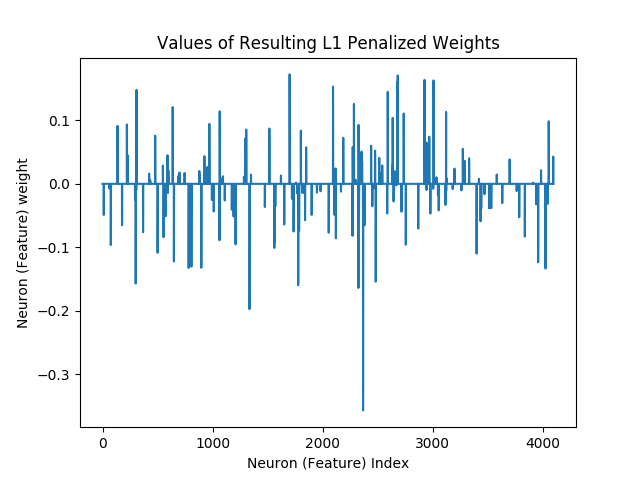}}
 \caption{Weights of 161 L1 neurons. Note multiple prominent positive and negative neurons.}
 \label{fig:final_weights}
\end{figure}

To control the sentiment of the music generated by our mLSTM, we find the
subset of neurons that contain the sentiment signal by exploring the weights 
of the trained logistic regression model. Since each of the 10 generative models derived 
from the 10 fold splits in Table \ref{tab:sent_anal} are themselves a full model, we use 
the model with the highest accuracy.
As shown in Figure 
\ref{fig:final_weights}, the logistic regression trained with regularization ``L1'' 
uses 161 neurons out of 4096. Unlike the results of Radford et al. \cite{radford_2017},
we don't have one single neuron that stores most of the sentiment signal. Instead,
we have many neurons contributing in a more balanced way. Therefore, we can't simply
change the values of one neuron to control the sentiment of the output music. 

We used a Genetic Algorithm (GA) to optimize the weights of the 
161 L1 neurons in order to lead our mLSTM to generate only positive
or negative pieces. Each individual in the population of this GA has $161$ 
real-valued genes representing a small noise to be added to the weights of the $161$ L1 neurons. The fitness of an individual is computed by (i) adding the genes of the individual to the weights (vector addition) of the $161$ L1 neurons of the generative mLSTM, (ii) 
generating $P$ pieces with this mLSTM, (iii) using the logistic regression model to predict 
these $P$ generated pieces and (iv) calculating the mean squared error of the $P$ 
predictions given a desired sentiment $s \in S = \{0, 1\}$.

The GA starts with a random population of size 100 where each gene of
each individual is an uniformly sampled random number $-2 \leq r \leq 2$.
For each generation, the GA (i) evaluates the current population, 
(ii) selects 100 parents via a roulette wheel with elitism, (iii) recombines the
parents (crossover) taking the average of their genes and (iv) mutates each
new recombined individual (new offspring) by randomly setting each gene to 
an uniformly sampled random number $-2 \leq r \leq 2$.  

We performed two independent executions of this GA,
one to optimize the mLSTM for generating positive pieces
and another one for negative pieces. Each execution optimized the
individuals during 100 epochs with crossover rate of 95\% and 
mutation rate of 10\%. To calculate the fitness of each individual, we generated $P$=30 pieces with 256 words each, starting with the symbol ``.'' (end of time step).
The optimization for positive and negative generation resulted in best individuals with
fitness $0.16$ and $0.33$, respectively. This means that if we
add the genes of the best individual of the final population to the weights of the generative mLSTM, we generate positive pieces
with 84\% accuracy and negative pieces with 67\% accuracy. 

\begin{figure}[!h]
 \centerline{\includegraphics[width=\columnwidth]{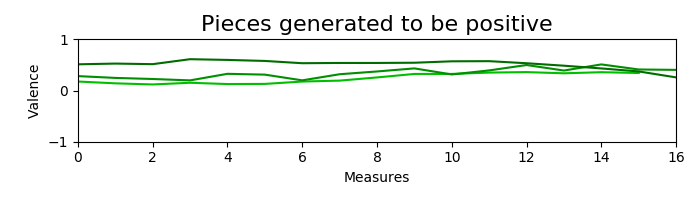}}
 \centerline{\includegraphics[width=\columnwidth]{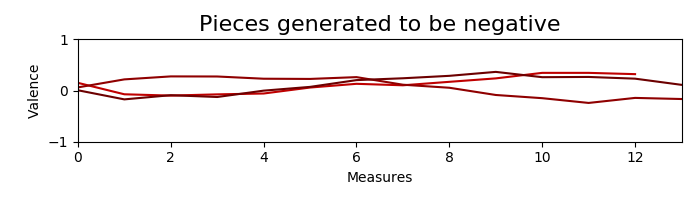}}
 \caption{Average valence of the 6 generated pieces, as determined by human annotators. 
 with least variance.}
 \label{fig:generated_eval}
\end{figure}

After these two optimization processes, the genes of the 
best final individual of the positive optimization were
added to the  weights of the 161 L1 neurons of the trained 
generative mLSTM. We then generated 30 pieces with 1000 
words starting with the symbol ``.'' (end of time step) and 
randomly selected 3 of them. The same process was repeated 
using the genes of the best final individual of the 
negative execution. We asked annotators to label this 6 
generated pieces via Amazon MTurk, using the the same 
methodology described in Section \ref{sec:data_collection}.
Figure \ref{fig:generated_eval} shows the average valence per measure of each of the generated pieces. 

We observe that the human annotators agreed that the three positive generated pieces are indeed positive. 
The generated negative pieces are more ambiguous, having both negative and positive measures.  However, as a whole the negative pieces have lower valence 
than the positive ones. This suggests that
the best negative individual (with fitness $0.33$) encountered by the GA 
wasn't good enough to control the mLSTM to generate complete negative pieces. Moreover, the challenge to optimize the L1 neurons suggests that there are more positive pieces than negative ones in the 3 shards used to train the generative mLSTM.




\section{Conclusion and Future Work}

This paper presented a generative mLSTM that can be controlled to generate symbolic music with a given sentiment. The mLSTM is controlled by optimizing the weights 
of specific neurons that are responsible for the sentiment signal. Such neurons are 
found plugging a Logistic Regression to the mLSTM and training the Logistic Regression 
to classify sentiment of symbolic music encoded with the mLSTM hidden states. We evaluated
this model both as a generator and as a sentiment classifier. Results showed that 
our model obtained good classification accuracy, outperforming a equivalent LSTM 
trained in a fully supervised way. Moreover, a user study showed that humans agree
that our model can generate positive and negative music, with the caveat that the 
negative pieces are more ambiguous. 

In the future, we plan to improve our model to generate less ambiguous negative pieces. Another
future work consists of expanding the model to generate music with a given
emotion (e.g. happy, sad, suspenseful, etc.) as well as with a given valence-arousal pair 
(real numbers). We also plan to use this model to compose soundtracks in real-time for 
oral storytelling experiences  \cite{padovani2017bardo}. 

\section{Acknowledgments}

We would like to thank Dr. Levi Lelis for the great feedback and Dr. Leonardo N. Ferreira for the 
support on the time series analysis. This research was supported by CNPq (200367/2015-3).  

\bibliography{ISMIRtemplate}

%
%
%
%

\end{document}